%% file: neurips_2024.tex
\documentclass{article}

\usepackage[final]{neurips_2024}

\usepackage[utf8]{inputenc} 
\usepackage[T1]{fontenc}    
\usepackage{hyperref}       
\usepackage{url}            
\usepackage{booktabs}       
\usepackage{amsfonts}       
\usepackage{nicefrac}       
\usepackage{microtype}      
\usepackage{xcolor}         

\usepackage{wrapfig}
\usepackage{makecell}
\usepackage{multirow}
\usepackage{pifont}
\usepackage{colortbl}
\definecolor{aliceblue}{rgb}{0.94, 0.97, 1.0}
\usepackage{array}
\usepackage{graphicx}
\usepackage[ruled, vlined]{algorithm2e}
\usepackage{hyperref}
\input{math_commands}

\DeclareMathOperator*{\Sim}{Sim}

\title{Dual-Personalizing Adapter for \\ Federated Foundation Models}

\author{%
  Yiyuan Yang \\
  Australian AI Institute, \\
  Faculty of Engineering \& IT\\
  University of Technology Sydney \\
  \texttt{Yiyuan.Yang-1@student.uts.edu.au} \\
   \And
  Guodong Long\\
  Australian AI Institute, \\
  Faculty of Engineering \& IT\\
  University of Technology Sydney \\
  \texttt{Guodong.Long@uts.edu.au} \\
   \AND
  Tao Shen\\
  Australian AI Institute, \\
  Faculty of Engineering \& IT\\
  University of Technology Sydney \\
  \texttt{Tao.Sheng@uts.edu.au} \\
   \And
  Jing Jiang\\
  Australian AI Institute, \\
  Faculty of Engineering \& IT\\
  University of Technology Sydney \\
  \texttt{Jing.Jiang@uts.edu.au} \\
   \And
  Michael Blumenstein\\
  Australian AI Institute, \\
  Faculty of Engineering \& IT\\
  University of Technology Sydney \\
  \texttt{Michael.Blumenstein@uts.edu.au} \\
}

\begin{document}

\maketitle

\begin{abstract}
  Recently, foundation models, particularly large language models (LLMs), have demonstrated an impressive ability to adapt to various tasks by fine-tuning diverse instruction data. Notably, federated foundation models (FedFM) emerge as a privacy preservation method to fine-tune models collaboratively under federated learning (FL) settings by leveraging many distributed datasets with non-IID data. To alleviate communication and computation overhead, parameter-efficient methods are introduced for efficiency, and some research adapted personalization methods to FedFM for better user preferences alignment. However, a critical gap in existing research is the neglect of test-time distribution shifts in real-world applications, and conventional methods for test-time distribution shifts in personalized FL are less effective for FedFM due to their failure to adapt to complex distribution shift scenarios and the requirement to train all parameters. 
  To bridge this gap, we refine the setting in FedFM, termed test-time personalization, which aims to learn personalized federated foundation models on clients while effectively handling test-time distribution shifts simultaneously. To address challenges in this setting, we explore a simple yet effective solution, a \textbf{Fed}erated \textbf{D}ual-\textbf{P}ersonalizing \textbf{A}dapter (FedDPA) architecture.
By co-working with a foundation model, a global adapter and a local adapter jointly tackle the test-time distribution shifts and client-specific personalization. 
Additionally, we introduce an instance-wise dynamic weighting mechanism that dynamically integrates the global and local adapters for each test instance during inference, facilitating effective test-time personalization. The effectiveness of the proposed method has been evaluated on benchmark datasets across different NLP tasks with released \href{https://github.com/Lydia-yang/FedDPA}{code}.
\end{abstract}

\section{Introduction}

Foundation models, especially the large language model (LLM) in natural language processing (NLP), have nearly exhausted public data sources for training. This necessitates alternative solutions to further improve these foundation models by leveraging private or protected data sources, such as business data in companies, smartphones, and so on. 
\textbf{Federated foundation models (FedFM)}\cite{zhuang2023foundation,yu2023federated} offer a promising solution by integrating federated learning (FL) frameworks to enhance the foundation models in a decentralized manner. 
Built upon existing Parameter-efficient fine-tuning (PEFT) methods \cite{xu2023parameter,hu2023llm,hu2021lora}, FedFM is a collaboratively fine-tuning framework that leverages private datasets with privacy preservation and avoiding overfitting from client-specific fine-tuning.


Test-time distribution shift in FL \cite{jiang2023test,tan2024heterogeneity} presents a significant challenge in practical scenarios, as clients may encounter unseen learning tasks during the testing and model inference phases.
For example, a client accustomed to writing emails in English may require translation assistance when working on a new project in Chinese. Therefore, it is imperative for the deployed machine learning model to be capable of tackling the test-time distribution shifts from the training data, and our paper addresses this critical issue of test-time distribution shifts within the FedFM scenario. Previous works in test-time FL\cite{tan2024heterogeneity,jiang2023test} predominantly utilize conventional deep learning models that are trained from scratch in federated settings. Recent FedFM methods mainly focus on addressing specific challenges related to data heterogeneity \cite{babakniya2023slora,jiang2023low} and communication overheads \cite{xu2023federated,sun2023fedbpt}. However, none of these methods have discussed test-time distribution shifts in FedFM scenarios.

To fill this gap, we propose a novel FedFM framework that is robust to client-specific alignment and test-time distribution shifts simultaneously. With the support of a foundation model with PEFT methods, we first refine the federated setting, termed \emph{test-time personalization}, which follows: 
1) each client needs to train a personalized model using its own data from a target task, and 2) during testing, each client's personalized model needs to be robust to tackle the receiving new tasks (unseen in training) with different distributions (test-time distribution shift).
Essentially, the proposed test-time personalization in FL could be simply viewed as an optimization task to seek a sweet trade-off between client-specific model personalization and model generalization to test data. 

For test-time personalization in FedFM, two primary challenges—test-time distribution shifts and personalization—necessitate learning tailored to distinct objectives, and the training cost of foundation models also represents a significant concern. To address these issues, we explore a simple yet effective method, dubbed \textbf{Fed}erated \textbf{D}ual-\textbf{P}ersonalizing \textbf{A}dapter (FedDPA), where each client learns a global adapter to learn generic knowledge from the aggregation for test-time tasks and maintains a local adapter for targeted ability personalization. During the inference phase, the local and global adapters are dynamically integrated to facilitate prediction, where an instance-wise dynamic weighting mechanism is proposed to autonomously adjudicate the proportional contribution of the local and global adapters for each test instance.
Experimental results demonstrate that our method achieves state-of-the-art performance on benchmarks and all data and code are released \footnote{https://github.com/Lydia-yang/FedDPA}. Our main contributions are summarized as follows:

\begin{itemize}
\item We are the first to explore the test-time distribution shifts problem in federated foundation models for practical application scenarios alignment.

\item We introduce a new method, namely dual-personalizing adapter, to emphasize learning both generic and personalized knowledge in the context of FedFM with test-time personalization.

\item We conduct an exhaustive analysis using heterogeneous FL benchmarks across diverse NLP tasks. The empirical outcomes reveal that our method attains state-of-the-art performance, underscoring its superior test-time personalization capabilities than existing methods.
\end{itemize}

\section{Related Work}

\subsection{ Adapter-based PEFT Methods} 

Given the substantial computational and storage burdens associated with directly fine-tuning foundation models, the community has shifted towards embracing parameter-efficient methods \cite{xu2023parameter}, with the adapter family \cite{hu2023llm} being a notable exemplar. According to different architectures, methods in the adapter family can be categorized into four types. The first one is prompt-based learning \cite{lester2021power,li2021prefix}, which is aimed at learning the continuous/soft prompt for discrete optimization. The second one is reparametrization-based methods \cite{hu2021lora,edalati2022krona}, achieving parameter efficiency by utilizing low-rank techniques to decompose the high-dimensional matrices. The third one is series Adapters \cite{houlsby2019parameter}, which introduce additional learnable modules in a sequential manner within specific sublayers. The last one is parallel Adapters \cite{he2021towards}, which focus on learning additional learnable modules in a parallel way with distinct sublayers. In this context, our exploration delves into the adapter-based PEFT methods of federated foundation models.

\subsection{Federated Foundation Models}
With the advent of foundation models, there has been a burgeoning interest  \cite{zhuang2023foundation,yu2023federated,ren2024advances,charles2024towards} in integrating these models within the FL setting. 
Particularly, in light of the inherent computation and communication cost, recent work \cite{kuang2023federatedscope,zhang2023fedpetuning,chen2024feddat} endeavors have delved deeper into integrating adapter-based parameter-efficient tuning (PEFT) methods with federated foundation models. 
Building upon this, a multitude of studies have emerged to navigate the challenges of incorporating federated foundation models with adapter-based PEFT methods.
The paper \cite{zhang2023towards} stands at the forefront, initiating the integration of instruction tuning within federated LLM frameworks.
Addressing data-related issues, the paper \cite{babakniya2023slora} introduced a data-driven initialization approach to mitigate the primary challenges associated with LoRA in highly heterogeneous data scenarios. In addition, the research presented in \cite{jiang2023low} proposed a method to annotate unlabeled client-side data by harnessing the prowess of large models to address data scarcity concerns.
To further optimize the communication and computational overheads associated with federated foundation models, the works \cite{xu2023federated,sun2023fedbpt,xu2024fwdllm} emphasize advancing gradient-free optimization methods suitable for devices with limited memory and computing power.
For personalization, paper \cite{yi2023fedlora} focused on designing a specific training paradigm for LoRA to achieve more effective personalization in visual model-heterogeneous scenarios. Diverging from these approaches, our work delves into the realm of personalization with adapters in federated foundation models, extending the scope of research in this area.

\subsection{Personalized Federated Learning}

To address the necessity of personalization for individual clients, personalized Federated Learning (PFL) \cite{tan2022towards}, which aims at training to cater to individual client preferences and needs, is proposed. Broadly, existing PFL methods can be categorized into two primary types: fine-tuning the global model for personalization or learning additional personalized models. Research works \cite{fallah2020personalized,collins2021exploiting} in the first category fine-tuned the whole or part of the global model with each client’s local dataset for personalization. While research works \cite{li2021ditto,li2021fedbn} in the second category is to learn the additional personalized layers or model through local aggregation. Nonetheless, a prevalent limitation among these PFL approaches is their concentrated focus on a specifically targeted task, often at the expense of performance when encountering test-time distribution shifts. 

To fill this gap, recent research has shifted focus towards exploring different test-time distribution shifts in PFL. In contrast to studies \cite{yoon2021federated,dong2022federated} in federated continual/incremental learning possessing ample annotated data from different distributions for training to address shifts, test-time FL focuses on handling distribution shifts during testing without the availability of annotated data for further training. One strand of research \cite{bao2024adaptive,xu2023joint} concentrates on addressing test-time distribution shifts that occur when new clients are introduced during the testing phase by module/prior adaptation.  
Another line of studies\cite{tan2024heterogeneity,jiang2023test} aims to tackle distribution shifts in existing clients during testing by aligning test features with existing features. Our paper falls into the second type and differs from previous work by exploring this challenge within the framework of foundation models, which are characterized by extensive parameter scales and more complex test-time distribution shifts.


\begin{figure}
  \centering
  \includegraphics[width=\textwidth]{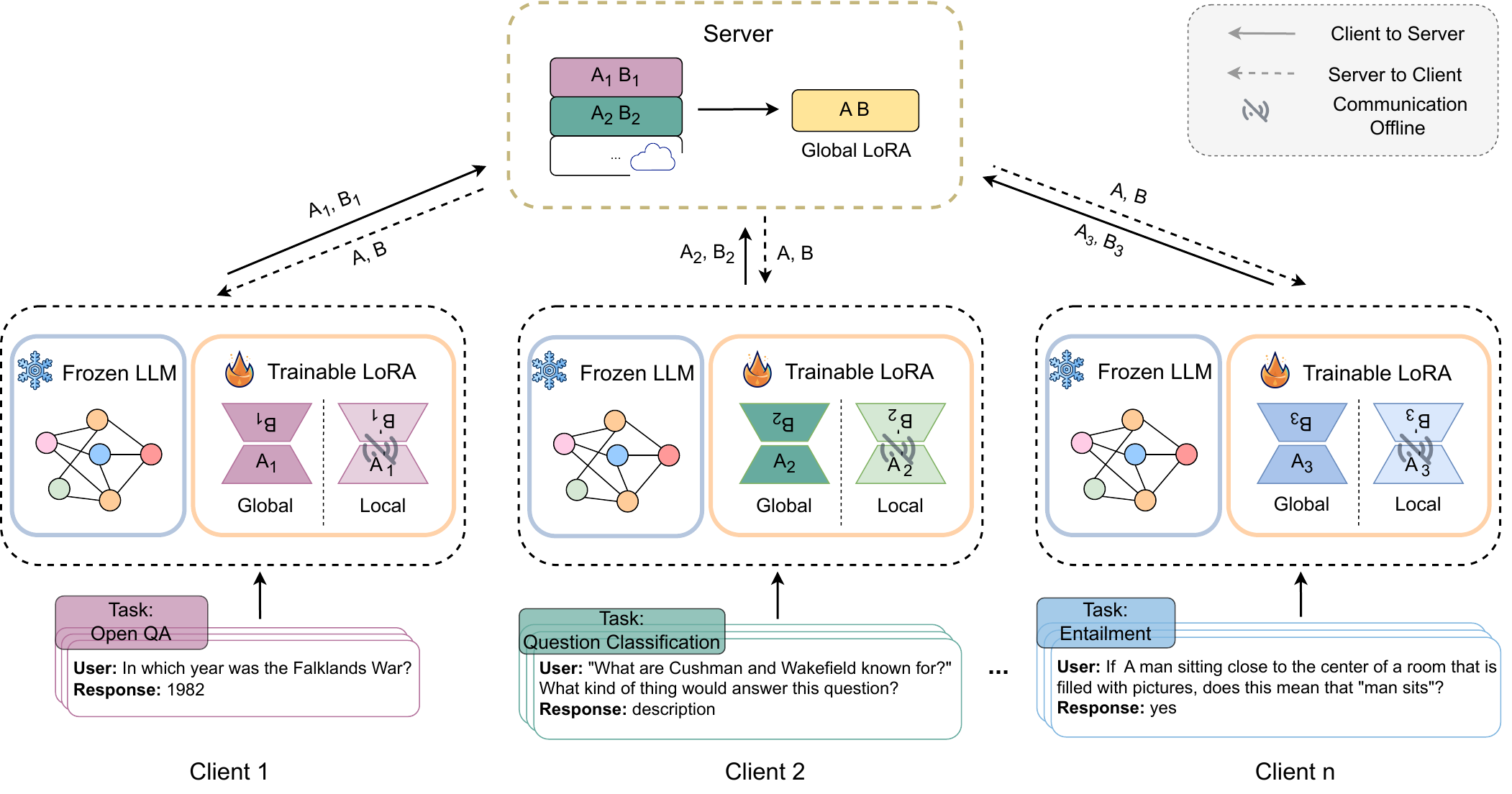}
  \caption{The overall framework of FedDPA. Each client contains a frozen LLM, a trainable global adapter (LoRA) and a trainable local adapter (LoRA) with a specific task, where the global adapter (LoRA) is for test-time tasks and the local adapter (LoRA) is for personalization. During the training, only the parameters of the global adapter (LoRA) are transmitted to the server for aggregation.}
  \label{fig:framework}
\end{figure}

\section{Problem Definition}

\subsection{Test-time Personalization in FedFM}
Considering $M$ clients in an FL system, each client possesses its distinct local training dataset $\sD^m_{train}$ and test dataset $\sD^m_{test}$, where $m$ indexes a client. One data pair in datasets is denoted as ${(\vx,\vy)}$, where $\vx$ is the input data and $\vy$ is its corresponding label.

In each client $m$, we will introduce the training and testing phases separately for our test-time personalization setting. In the training phase, model training utilizes solely the local dataset $\sD_{train}^m$, which is derived from the distribution $P_s^m$. 
While in the testing phase, the test dataset $\sD_{test}^m$ comprises two components: the test set $\sD_s^m$ driven from the same distribution $P_s^m$ as training data, and additional test sets $\sD_t^m$ under data distribution shifts $P_s^m(\vx,\vy) \ne P_t^m(\vx,\vy)$. Therefore, the test dataset is $\sD_{test}^m=\sD_s^m \cup\sD_t^m$, and we call these datasets $\sD_t^m$ as test-time datasets. Unlike previous works\cite{tan2024heterogeneity,jiang2023test} in test-time FL concentrating on either feature-level shifts $P_s^m(\vx) \ne P_t^m(\vx)$ or label shifts $P_s^m(\vy) \ne P_t^m(\vy)$, we investigate a more complex scenario where various distribution shifts, including semantic shifts, domain shifts and others, exist simultaneously. This is aligned with the practical application of foundation models, which often encounter testing data originating from diverse domains, backgrounds, or populations.

Therefore, the objective of the model in each client should not only perform well on the test set $\sD_s^m$ (refer to personalization) but also have comparable results on the test-time dataset $\sD_t^m$ (refer to test-time performance). This objective is consistent with the practical scenarios, since users primarily focus on the abilities they often utilize (abundant data available for training) and occasionally also introduce new tasks (limited to test data).

\subsection{Challenges} 
The test-time personalization setting raises two pivotal considerations: \emph{personalization} and \emph{test-time distribution shifts}. Personalization is the primary focus, followed by optimizing test-time tasks. Our proposed method introduced in section~\ref{method} is designed to achieve personalization within FedFM while ensuring comparable results for test-time tasks, and its vital intuition is illustrated below.

Considering a foundation model, it comprises a main body $f(\vtheta)$, which holds most of the parameters and processes input $\vx$ to produce output features $\vh=f(\vx;\vtheta)$. Additionally, there is a tail $g(\vtheta_t)$ that maps these features to the output space (e.g., vocabulary), resulting in the predicted result $\hat{\vy}=g(\vh;\vtheta_t)$. Typically, the focus in tuning and adaptations primarily lies on the main body $f(\vtheta)$ because the tail $g(\vtheta_t)$, usually a linear function, remains unchanged (frozen) during tuning \cite{hu2023llm}.  

\paragraph{Discordance between Personalization and Test-time Tasks.}
The key to addressing test-time distribution shifts lies in learning generic features universally applicable across disparate distributions \cite{arjovsky2020out}. That is, learning a foundation model $f(\vtheta)$ to satisfy $P_s(f(\vx;\vtheta),\vy) = P_t(f(\vx;\vtheta),\vy)$ although $P_s(\vx,\vy) \ne P_t(\vx,\vy)$. FL is a methodology designed to learn generic features across diverse non-IID data (different distributions) through aggregation algorithms \cite{charles2024towards,tan2024heterogeneity}. Therefore, we tailor FL training for addressing test-time tasks with the objective $\min_\vtheta\Ls_{P_{all}}({\vtheta})$, where $\Ls_{P_{all}}$ represents the loss function designed for learning generic features towards all clients' distributions $P_{all}$.
However, personalization focuses on aligning the model with the specific distribution $P_s$, which means learning a foundation model $f(\vtheta)$ with the objective $\min_\vtheta\Ls_{P_s}({\vtheta})$, where $\Ls_{P_s}$ represents the loss function designed for learning personalized features towards the specific distribution. Therefore, the discordance between specific distribution alignment for personalization and generic feature learning for test-time tasks leads to inconsistent optimization objectives. 

The above analysis motivates a dual model strategy—one model for test-time tasks and one model for personalization—to realize test-time personalization in FedFM. This strategy, together with our other techniques presented below for FedFM scenarios, constitutes the foundation of our method.

\section{Proposed Method}

\label{method}
To align with the application scenarios, we consider the test-time personalization setting in FedFM. 
Following a similar assumption from a Mixture of Experts \cite{masoudnia2014mixture}, \emph{any test-time task (distributions) to a client can be approximated as a mixture of training tasks seen by other clients in the federated learning system}. Therefore, each client primarily personalizes its model based on its local training task, while also tackling unseen test-time tasks by leveraging insights gained from other clients in the federated learning system. Discussions of other scenarios can be found in Appendix~\ref{limit}.

In test-time personalization, test-time distribution shifts and personalization are two main issues that need to be addressed, and their optimization objectives toward different distributions are inconsistent. To address these challenges and consider the efficient learning of FedFM, we propose a \textbf{Fed}erated \textbf{D}ual-\textbf{P}ersonalizing \textbf{A}dapter (FedDPA) system for each client, as shown in Fig~\ref{fig:framework}. During training, a global adapter is employed to acquire generic features by FL training for test-time tasks (Sec.~\ref{global_adapter}). Meanwhile, to address personalization, a local adapter is maintained locally to align with the client's specific distribution, and leverages generic knowledge from the global adapter for faster learning (Sec.~\ref{local_adapter}). During the inference, the learned global and local adapters are dynamically combined using a weight generated by the instance-wise dynamic weighting mechanism for each input test instance, realizing test-time personalization (Sec.~\ref{auto}). 

\paragraph{The Overall Objective.} Considering the computation and communication cost of FedFM, we utilize the adapter-based PEFT methods, which only learn a small part of parameters $\Delta\vtheta$ while keeping most of the parameters $\vtheta$ frozen. Our proposed FedDPA is to learn the global adapter $\Delta\vtheta_g$ and local adapters $\Delta\vtheta_l^m$ \emph{simultaneously} across $M$ client to realize test-time personalization,
\begin{equation}
  \begin{aligned}
     &\min_{\Delta\vtheta_g,\{\Delta\vtheta_l^m\}} &\sum_{m=1}^M [ \Ls_{(\vx,\vy)\sim P_s^m} (\vtheta;\Delta\vtheta_g;\Delta\vtheta_l^m)] \\
     &s.t. & \Delta\vtheta_g^* \in \argmin_{\Delta\vtheta_g} \Ls_{(\vx,\vy)\sim P_{all}} (\vtheta;\Delta\vtheta_g;\Delta\vtheta_l^m) 
  \end{aligned}
\end{equation}

where the first part is a standard personalized FL loss to find optimal personalized models by minimizing the sum of loss on local training tasks $\Ls_{(\vx,\vy)\sim P_s^m}(.)$, and the second part is a constraint term to seek an optimal solution by minimizing the test-time loss $\Ls_{(\vx,\vy)\sim P_{all}(.)}$. Because we assume that the test-time task is unseen to a client but observed by other clients, the test-time loss can be estimated using the Empirical Risk Minimization of all client's training tasks $\min_{\Delta\vtheta_g}\sum^M_{m=1} r_m \Ls_m (\vtheta;\Delta\vtheta_g)$, where $P_{all}$ denotes all distributions of tasks in all clients, $r_m$ denotes each client's weight for aggregation (e.g., in FedAvg, $r_m$ is the proportion of each client's data number to all clients' data number) and $\Ls_m$ denotes the loss for each client over its local training dataset. Since the above objective cannot be solved directly, \emph{we propose to alternatively learn the global and local adapters in a sequential manner (FedDPA-F with local adapter fine-tuning) or iterative manner (FedDPA-T with local adapter training)}. Detailed algorithms of these two methods are in Appendix~\ref{algorihtms}.


\paragraph{Remark.} To simplify the illustration, we use LLM as the backbone and adopt LoRA \cite{hu2021lora} as the adapter-based PEFT method in our framework. The overall framework is easy to adapt to other types of backbone and other adapter-based PEFT methods. LoRA decomposes the training weight into a frozen weight $\vtheta$, and a trainable weight derived by the multiplication of two low-rank weights $\Delta \vtheta = \Delta\vtheta^b\Delta\vtheta^a$. The data heterogeneity in FL with LLM primarily manifests as distribution shifts across various NLP tasks among different clients, driven by diverse backgrounds, topics, and other contextual factors, and local loss $\Ls_m$ for all NLP tasks is a standard language modeling objective \cite{brown2020language}.  

\subsection{Generic Learning of Global Model}
\label{global_adapter}
Addressing test-time distribution shifts requires the acquisition of generic knowledge that is applicable across various distributions \cite{arjovsky2020out}. The conventional federated learning process is inherently designed to aggregate this generic knowledge among different non-IID data. Consequently, we utilize the adapter trained within the FL context as the global adapter for addressing test-time tasks. To further enhance generic learning, our model aggregation strategy is based on the client number rather than the number of data by considering the potential biases stemming from different numbers of tasks.

At each client, there consists of a frozen LLM model $f(\vx;\vtheta)$ with a global lightweight global adapter (LoRA) $\Delta \vtheta_g = \Delta \vtheta^b_g\Delta\vtheta^a_g$. This global adapter is used for aggregation by sending to the server. Notably, the server's role is limited to computing the aggregated adapter $\Delta \vtheta_g$, thus obviating the need for maintaining a large-scale model. Similar to the standard FL process, for each client $m$, the adapter weight $\Delta \vtheta_g^m$ is learned locally and sent to the server. Upon receipt of the adapter weights from all clients, the server employs FedAvg \cite{mcmahan2017communication} to aggregate them and sends $\Delta \Bar{\vtheta_g}$ back to each client as their initialized parameter in a new round. It can be formulated as:
\begin{equation}
\label{eq-global}
    \textbf{Server: } \Delta\Bar{\vtheta_g}= \sum^M_{m=1} \frac{1}{M}\Delta\vtheta_g^m, \quad \textbf{Client: } \Delta\vtheta_g^m =\text{arg}\min_{\Delta \vtheta_g}\Ls_m(\vtheta;\Delta\vtheta_g), \text{ initialized with $\Delta\Bar{\vtheta_g}$}
\end{equation}
\paragraph{Remark.} Other federated algorithms like FedProx \cite{li2020federated} can also be applied with LoRA tuning of this global model learning (in Appendix~\ref{adapt}). In this paper, we just take FedAvg as an example.

\subsection{Personalization of Local Model}
\label{local_adapter}
\begin{figure}[h]
\begin{center}
\includegraphics[width=0.95\textwidth]{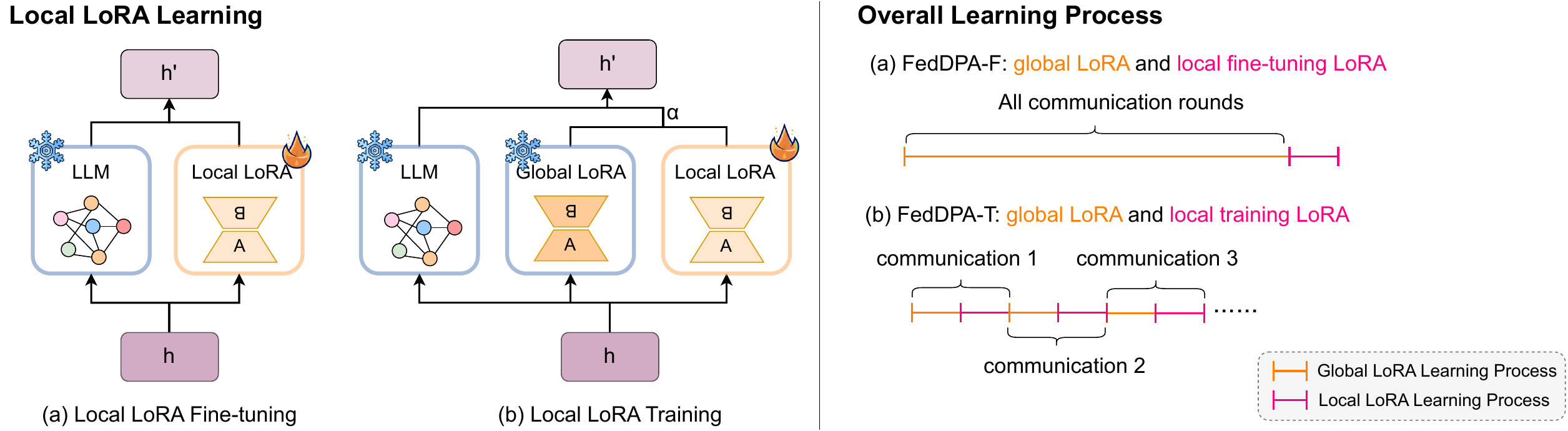} 
\end{center}
\caption{Frameworks of two personalized methods for local adapter (LoRA) are shown on the left, with their overall learning processes on the right.}
\label{fig:local}
\end{figure}

The previously developed global model, which focuses on acquiring generic features across diverse datasets, faces challenges with personalization due to inconsistent optimization objectives. To address this, we integrate a local adapter to better align with each client's specific distribution. We explore two methods as shown in Fig~\ref{fig:local}, 1) Learning sequentially: after global adapter training, the local adapter is initialized by the learned global adapter and directly fine-tuned; 2) Learning iteratively: during each communication round of global adapter training, the local adapter is re-initialized from its last state, fine-tuned alongside the frozen global adapter, and maintained locally without communication. 

To be more specific, a local adapter (LoRA) $\Delta \vtheta_l = \Delta\vtheta^b_l\Delta\vtheta^a_l$ is introduced. Thus, each client contains three components: a frozen LLM $\vtheta$, a global adapter (LoRA) $\Delta\vtheta_g$ and a local adapter (LoRA) $\Delta \vtheta^m_l$. As delineated in Fig~\ref{fig:local} (a), for the first method, after global training, the local adapter is first initialized by the global adapter denoted as $\Delta \vtheta^m_l=\Delta\vtheta_g$, then fine-tuned on local data to get the final local adapter. As shown in Fig~\ref{fig:local} (b), for the second method, during each communication round in training for each adapter layer, upon receiving an input $\vh$, it simultaneously traverses the frozen LLM, the frozen global adapter and the local adapter. The process entails an initial fusion of the outputs from both the local and global adapters with a predefined weighting factor of $\alpha$, followed by integration with the output of the LLM to yield the final result $\vh^{'} =\vtheta \vh+((1-\alpha) \cdot\Delta\vtheta_g \vh+\alpha \cdot\Delta\vtheta^m_l\vh)$. Therefore, the learning of the local adapter $\Delta\vtheta^m_l$ for these two methods can be unified as:
\begin{equation}
\label{eq-local}
\Delta\vtheta^m_l=\text{arg}\min_{\Delta\vtheta^m_l}\Ls_m(\vtheta;\Delta\vtheta_g;\Delta\vtheta^m_l), \quad \text{initialized with $\Delta \vtheta_g$ or previous $\Delta\vtheta_l^m$}
\end{equation}


\subsection{LLM-enhanced Instance-wise Dynamic Weighting Mechanism}
\label{auto}
  
As discussed in previous test-time FL methods \cite{jiang2023test}, a dynamic combination of global components and personalized components can improve generalization while reducing the cost of hyper-parameter tuning in the deployment stage. Considering the disparate data distributions that characterize test-time tasks and local tasks and the wealth of training instances of local tasks available to each client, we propose an \emph{instance-wise dynamic weighting mechanism} to calculate the similarity between the input instance and local instances, using this metric to determine the appropriate weight balance for the global and local adapter combination. To facilitate this, the representation of each input instance is essential. Leveraging the robust capability of pre-trained LLMs to abstract input sentences, we utilize the hidden states from the final layer of the LLM as the representation. Given that the LLM is decoder-based, with tokens attending only to preceding tokens, the embedding of the final token is considered representative of the entire input for similarity evaluation. Furthermore, to enhance the representation quality, the global adapter, which embodies generic knowledge, is incorporated into this embedding process. 

More specifically, during the inference stage, for each input instance $\vx$ in a client, we randomly sample $S$ instances $\{\vx_0,\vx_1,...,\vx_s\}$ from the local training dataset. These instances are then fed into the LLM, augmented with the global adapter, to obtain the last token's embeddings from the final layer, denoted as $\vw_x$ and $\{\vw_{x_0},\vw_{x_1},...,\vw_{x_s}\}$ respectively. Subsequently, we calculate the similarity between the input representation $\vw_x$ and each sampled local representation in $\{\vw_{x_0},\vw_{x_1},...,\vw_{x_s}\}$, resulting in a score range of $[0,1]$. Finally, we average all scores to obtain the final result, represented as $\alpha_{t}=\lambda\cdot\sum_{i=0}^S \frac{1}{S} \Sim(\vw_x,\vw_{x_i})$, where $\Sim$ represents the function to calculate the similarity, and $\lambda$ is a scale factor in $(0,1]$ to restrict the maximum similarity score (especially for FedDPA-T).

Through this method, the balancing of weights between the global and local adapters is dynamically adjusted for each test instance, ensuring the model not only tailors to the individual client's specific needs but also benefits from the aggregated model's generic knowledge across test-time tasks.

\section{Experiment}
\label{section:exp}
\subsection{Experiment Setting}
\paragraph{Datasets.} We construct two federated datasets from Flan \cite{wei2021finetuned}, which is a collection of various NLP tasks from over 60 datasets for instruction turning. 
In order to be better suitable for FL settings, we randomly select 8 NLP tasks from different datasets for each federated dataset and downsample the original datasets with more details in Appendix~\ref{dataset}.
ROGUE-1 is taken as a metric. 



\paragraph{Baselines and Implementation.} We compare our methods with four baselines based on the same model architecture: centralized model, Local-finetuned model, FedIT \cite{zhang2023towards} and FedLoRA \cite{yi2023fedlora}. The centralized model is trained on all data of tasks in one center. The local-finetuned model infers that only local data are used to train the model without any communication with other clients or the server. 

We distribute data between clients based on the NLP task for data heterogeneity, where different NLP tasks generated from different contextual factors inherently suffer from various complex distribution shifts. Since we select 8 NLP tasks, corresponding to $M=8$ clients.For each client, the local task serves as the primary focus for personalization, while the tasks from other clients are taken as test-time tasks. To better evaluate the effectiveness of methods, we assume that all clients are activated for every communication round and set the communication round $K=20$. The alpaca-LoRA\footnote{https://github.com/tloen/alpaca-lora} is adapted as the base model initialized with LLaMA-7B.\footnote{https://huggingface.co/huggyllama/llama-7b} 
The updating weight of local LoRA training (FedDPA-T) is $\alpha=0.5$ ($\lambda=0.5$) for federated dataset 1 and $\alpha=0.3$ ($\lambda=0.3$) for federated dataset 2. 
We set $S=5$ and choose cosine similarity for instance-wise dynamic weighting mechanism. More details are in Appendix~\ref{baselines}.
\definecolor{mygray}{gray}{0.95}
\begin{table}[h]
    \small
    \setlength{\tabcolsep}{4pt}
    \caption{Personalization and test-time personalization results of different models on federated dataset 1. FedDPA-F represents the model with the local fine-tuning adapter and FedDPA-T represents the model with the local training adapter. Linguistic represents the linguistic acceptability task, Word Dis represents the word disambiguation task, and Question CLS represents question classification task.}
    \centering
    \begin{tabular}{l|cccccccc|c}
        \toprule[1.25pt]
        \multirow{2}{*}{Methods}  & \multicolumn{9}{c}{\bf Federated Dataset 1} \\
        &\multicolumn{1}{l}{\makecell[c]{Para \\ -phrase}}  &\multicolumn{1}{l}{\makecell[c]{Entail \\ -ment}}  &\multicolumn{1}{l}{\makecell[c]{Structure \\ to Text}} &\multicolumn{1}{l}{\makecell[c]{Text For \\ -matting}} &\multicolumn{1}{l}{\makecell[c]{Linguistic \\ Acc}}  &\multicolumn{1}{l}{\makecell[c]{Word \\ Dis}} &\multicolumn{1}{l}{\makecell[c]{Core \\ -ference}} &\multicolumn{1}{l}{\makecell[c]{Question \\ CLS}} &\multicolumn{1}{l}{\makecell[c]{Average}}
\\ \hline 
\rowcolor{mygray}\multicolumn{10}{l}{ \emph{Personalization}} \\
Centralized   &77.00  &82.00  &72.58  &96.59  &70.50 &63.50 &77.59 &89.00 &78.60 \\
FedIT         &69.00  &83.00  &71.25  &96.32  &71.50 &62.50 &75.43 &91.50 &77.50 \\
FedLoRA       &77.50  &84.00  &71.49  &96.69  &73.50 &\textbf{65.00} &75.27 &92.00 &79.43 \\
Local-finetuned  &74.50  &80.00  &\textbf{73.71}  &\textbf{97.36}  &\textbf{75.00} &54.50 &68.55 &89.50 &76.64 \\
\hline
FedDPA-F &79.00 &\textbf{84.50}  &72.06  &96.90  &72.00 &\textbf{65.00} &73.86 &92.50 &79.48 \\
\rowcolor{aliceblue!60} FedDPA-T     &\textbf{80.50}  &\textbf{84.50}  &72.79  &96.51  &73.50 &62.00 &\textbf{77.93} &\textbf{94.00} &\textbf{80.22}\\

\hline 
 \rowcolor{mygray}\multicolumn{10}{l}{ \emph{Test-Time Personalization}} \\
Local-finetuned   &48.99  &47.24  &27.53  &22.66  &48.86 &49.07 &46.45 &52.09 &42.86 \\
FedLoRA       &75.56  &76.55  &75.21 &74.94  &76.16 &74.64 &74.99 &76.97 &75.63 \\
\hline 
\rowcolor{aliceblue!60} FedDPA-F &\textbf{78.10} &\textbf{77.36}  &\textbf{77.18}  &\textbf{76.98}  &\textbf{77.11} &\textbf{76.23} &\textbf{76.84} &\textbf{77.19} &\textbf{77.12} \\
FedDPA-T     &76.20  &75.51  &76.19  &75.63  &74.86 &74.60 &74.77 &75.96 &75.47\\
        \bottomrule[1.25pt]
    \end{tabular}
    \label{table-best}
\end{table}
\begin{table}
    \small
    \setlength{\tabcolsep}{4pt}
    \caption{Personalization and test-time personalization results of different models on federated dataset 2. FedDPA-F represents the model with the local fine-tuning adapter and FedDPA-T represents the model with the local training adapter. Reading Com represents the reading comprehension task.}
    \label{table-data2}
    \centering
    \begin{tabular}{l|cccccccc|c}
        \toprule[1.25pt]
        \multirow{2}{*}{Methods}  & \multicolumn{9}{c}{\bf Federated Dataset 2} \\
        &\multicolumn{1}{l}{\makecell[c]{Para \\ -phrase}}  &\multicolumn{1}{l}{\makecell[c]{Common \\ -sense}}  &\multicolumn{1}{l}{\makecell[c]{Entail \\ -ment}} &\multicolumn{1}{l}{\makecell[c]{Text For \\ -matting}} &\multicolumn{1}{l}{\makecell[c]{Summari \\ -zation}}  &\multicolumn{1}{l}{\makecell[c]{Reading \\ Com}} &\multicolumn{1}{l}{\makecell[c]{Senti \\ -ment}} &\multicolumn{1}{l}{\makecell[c]{Open \\ QA}} &\multicolumn{1}{l}{\makecell[c]{Average}}
\\ \hline 
\rowcolor{mygray}\multicolumn{10}{l}{ \emph{Personalization}} \\
Centralized   &87.00  &64.67  &77.00  &90.65  &29.12 &76.00 &72.50 &76.17 &71.64 \\
FedIT         &86.00  &63.13  &79.00  &89.80  &30.36 &75.50 &72.00 &81.06 &72.07 \\
FedLoRA       &87.00  &64.12  &\textbf{84.50}  &89.52  &27.13 &76.50 &73.50 &79.62 &72.74 \\
Local-finetuned  &75.00 &53.51  &81.00  &91.28  &27.51 &69.00 &72.50 &79.31 &68.64 \\
\hline
FedDPA-F &88.00 &64.80  &84.25  &89.82  &29.58 &78.50 &72.00 &80.89 &73.48 \\
\rowcolor{aliceblue!60} FedDPA-T     &\textbf{90.50}  &\textbf{70.54}  &82.00  &\textbf{91.81}  &\textbf{30.75} &\textbf{81.00} &\textbf{75.00} &\textbf{91.07} &\textbf{75.33}\\

\hline 
 \rowcolor{mygray}\multicolumn{10}{l}{ \emph{Test-Time Personalization}} \\
Local-finetuned   &48.21  &49.07  &49.75  &21.86  &17.35 &48.57 &44.04 &48.19 &40.88 \\
FedLoRA       &69.60  &71.64  &71.09 &71.28  &65.63 &68.89 &70.32 &70.44 &69.86 \\
\hline 
\rowcolor{aliceblue!60} FedDPA-F &\textbf{71.64} &72.28  &\textbf{72.42}  &72.39  &\textbf{71.12} &\textbf{70.46} &\textbf{71.00} &\textbf{71.82} &\textbf{71.64} \\
FedDPA-T     &71.63  &\textbf{72.66}  &71.20  &\textbf{72.58}  &70.58 &69.21 &70.67 &71.62 &71.27\\
        \bottomrule[1.25pt]
    \end{tabular}
\end{table}

\subsection{Main Results}
We compare FedDPA with other baselines on two main evaluation facets: \emph{personalization} (scores on targeted local tasks) and \emph{test-time personalization} (average scores on all tasks including test-time tasks). 
As evidenced in Table~\ref{table-best} and Table~\ref{table-data2}, our proposed dual-personalizing adapter methods (both fine-tuning and training) exhibit superior performance in personalization compared to other baseline models, which demonstrates the effectiveness of local adapter maintenance for enhancing performance on the targeted local task. For test-time personalization, the FedDPA-F method stands out as the most effective among all personalized models, which suggests that incorporating learning from the global adapter can be instrumental in adapting to test-time distribution shifts for a more comprehensive model achievement. Additionally, given that the global adapter aggregated on different distributions matin certain generalization capabilities, the local adapter of FedDPA-F has better generalization performance than that of FedDPA-T, which leads to better performance on most test-time tasks. More importantly, it is noteworthy that while centralized or global models may yield higher average performances across all tasks, they fall short in excelling at specific tasks for personalization, aligning with the conclusions of the previous study \cite{wang2023far}.

\section{Analysis}
\subsection{Convergence Analysis}
We present the convergence analysis of our methods in Figure~\ref{fig:round}. Figure~\ref{fig:round} (a) compares our methods with other baselines for personalization, with the results showcasing the average performance on target local tasks across all clients. Notably, our methods exhibit a more rapid convergence compared to FedIT and achieve notable performance enhancements after five communication rounds. Despite sharing similar trends with FedLoRA, our approaches, particularly the FedDPA-T, ultimately outperform in personalization. For a more granular insight into test-time personalization convergence, Figure~\ref{fig:round} (b) compares average performance on all tasks, including each client's targeted local and test-time tasks. The results substantiate that our approaches demonstrate faster convergence rates, further bolstering the efficacy of our methods. 

\begin{minipage}{\textwidth}
\begin{minipage}[t]{0.48\textwidth}
    \setlength{\tabcolsep}{4pt}
    \small
    \makeatletter\def\@captype{table}
    \caption{Ablation study of instance-wise dynamic weighting mechanism (Auto). P represents personalization, and TTP represents test-time personalization.}
    \centering
    \begin{tabular}{ll|ll|ll}
        \toprule[1.25pt]
        \multirow{2}{*}{{Methods}} &\multirow{2}{*}{{Auto}} &\multicolumn{2}{c}{\bf Fed Dataset 1}  &\multicolumn{2}{c}{\bf Fed Dataset 2} \\
        & & \multicolumn{1}{c}{P} & \multicolumn{1}{c}{TTP} & \multicolumn{1}{c}{P} & \multicolumn{1}{c}{TTP} 
\\ \hline 
\multirow{2}{*}{\makecell[c]{FedDPA-F}}&\ding{55} &79.06 &76.97 &73.17 & \textbf{71.70}\\
&\ding{51} &\textbf{79.48} &\textbf{77.12} &\textbf{73.48} &71.64\\
\hline
\multirow{2}{*}{\makecell[c]{FedDPA-T}}&\ding{55} &79.57 &60.06 &73.75 &63.57\\
&\ding{51} &\textbf{80.22} &\textbf{75.47} &\textbf{75.33} &\textbf{71.27}\\
      
        \bottomrule[1.25pt]
    \end{tabular}
    \label{table-auto}

\end{minipage}
\hfill
\begin{minipage}[t]{0.48\textwidth}
    \setlength{\tabcolsep}{4pt}
    \small
    \makeatletter\def\@captype{table}
     \caption{Ablation study of updating weight. P represents personalization, and TTP represents test-time personalization.}
    \centering
    \begin{tabular}{ll|ll|ll}
        \toprule[1.25pt]
        \multirow{2}{*}{{Methods}} &\multirow{2}{*}{\textbf{ $\alpha$}} &\multicolumn{2}{c}{\bf Fed Dataset 1}  &\multicolumn{2}{c}{\bf Fed Dataset 2} \\
        & & \multicolumn{1}{c}{P} & \multicolumn{1}{c}{TTP} & \multicolumn{1}{c}{P} & \multicolumn{1}{c}{TTP} 
\\ \hline 
\multirow{3}{*}{\makecell[c]{FedDPA-T}}&\multicolumn{1}{l|}{0.3} &79.69 &\textbf{75.85} &\textbf{75.33} &\textbf{71.27}\\
&\multicolumn{1}{l|}{0.5} &\textbf{80.22} &75.47 &74.10 &70.72\\
&\multicolumn{1}{l|}{0.7} &79.88 &75.01 &74.04 &69.95\\
      
        \bottomrule[1.25pt]
    \end{tabular}
    \label{table-weight}
\end{minipage}
\\
\begin{minipage}[b]{0.48\textwidth}
\makeatletter\def\@captype{figure}
\includegraphics[width=\linewidth]{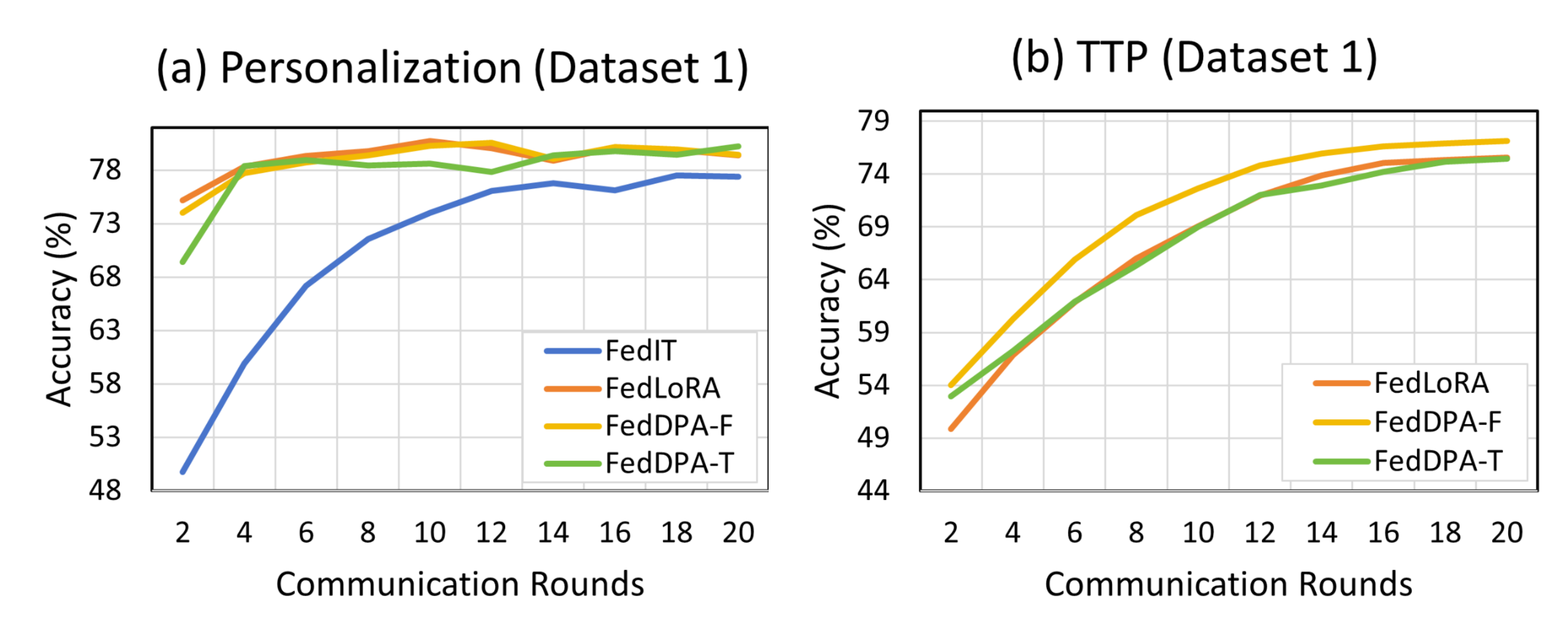} 
\caption{Average accuracy varies as communication rounds.}
\label{fig:round}
\end{minipage}
\hfill
\begin{minipage}[b]{0.48\textwidth}
\makeatletter\def\@captype{figure}
 \includegraphics[width=\linewidth]{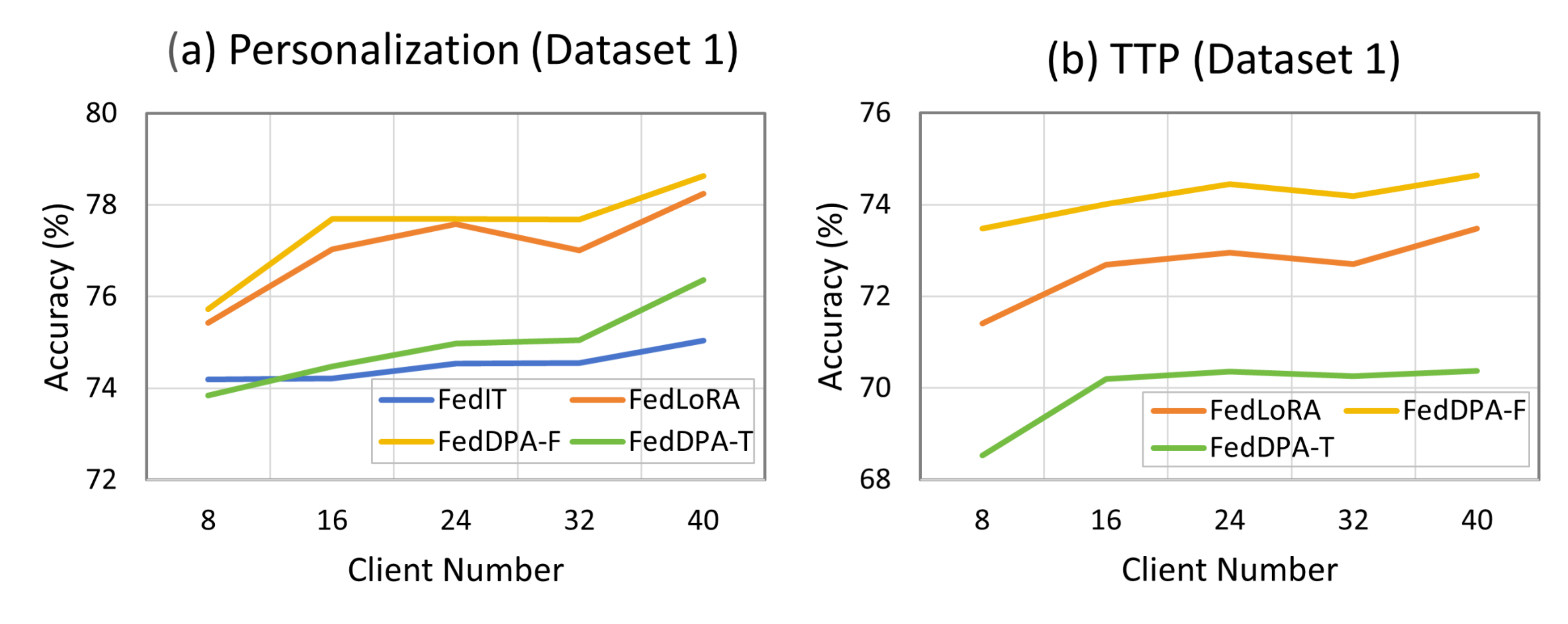} 
    \caption{Average accuracy varies as different client participation numbers.}
    \label{fig:clientnum}
\end{minipage}
\end{minipage}

\subsection{Ablation Study}
\label{section:ablation}
\paragraph{Impact of Instance-Wise Dynamic Weighting Mechanism.} To explore the impact of the instance-wise dynamic weighting mechanism, we implemented experiments with FedDPA methods on different datasets. As shown in Table~\ref{table-auto}, the incorporation of an instance-wise dynamic weighting mechanism contributes significantly to enhancing performance in both personalization and test-time personalization scenarios. More ablation studies are in Appendix~\ref{auto-exp}.

\paragraph{Impact of Updating Weight $\alpha$.} In this study, we investigated the influence of the updating weight $\alpha$ during FedDPA-T training with its value $\alpha \in \{03, 0.5, 0.7\}$. 
As can be seen in Table~\ref{table-weight}, for test-time personalization, increasing updating weight $\alpha$ will decrease the performance due to the increased proportion of the local adapter in the model, while for personalization, different updating weights $\alpha$ are required for different datasets to achieve their optimal results.

\paragraph{Impact of Client Number.} To better align with the FL setting in practical application, we scaled up clients to 40 and implemented experiments with sample rate $\{0.2,0.4,0.6,0.8,1\}$. 
For each communication round, the server will select clients from each task based on the sample rate (more details in Appendix~\ref{client_num}). 
As shown in Figure~\ref{fig:clientnum}, as the client participant rates increase, model accuracy also increases as more participating clients provide more data for knowledge learning. Besides, FedDPA-F outperforms all baselines, whereas FedDPA-T exhibits somewhat inferior performance, potentially due to overfitting issues when handling a small dataset. 

More experiments and analyses of scalability and efficiency can be found in Appendix~\ref{scale} and~\ref{computation}.

\section{Conclusion}
Federated Foundation Model (FedFM) is a promising direction to enhance existing Foundation Models, e.g. LLM, by leveraging private data sources. Test-time distribution shift is a critically important problem to ensure the practicability of the FedFM system. This work is the first to propose the test-time FedFM setting. To tackle this challenging scenario, we propose a novel dual-personalizing adapter for the FedFM framework. The method is evaluated on public NLP tasks that are adapted to mimic the test-time FedFM setting. This work is the first step towards this direction. We focus on defining a new learning scenario, proposing a basic learning framework, and setting up the benchmark datasets. Our future works will be in two directions: the first is to rethink this problem from a theoretical perspective, and the second is to enhance the benchmark setting with more datasets in real applications. 

\bibliographystyle{plain}
\bibliography{neurips_2024}







\newpage

\clearpage
\appendix
\section*{Appendix}
\section{Implementation Details}
\label{exp}
\subsection{Datasets}
\label{dataset}
In this paper, we have developed two federated datasets derived from the Flan \cite{wei2021finetuned}, and details to construct these datasets are elucidated in this section. Flan encompasses a diverse array of NLP tasks, each comprising multiple datasets. These tasks, generated from different contextual factors, inherently experience various complex distribution shifts. To align with FL settings, we employed a stratified selection process, randomly choosing one dataset from each of the eight distinct tasks from Flan to form each federated dataset. In addition, to simulate client local data scarcity \cite{mcmahan2017communication}, we implemented a downsampling strategy, reducing the size of each selected local dataset to 300 training instances and 200 testing instances. Consequently, each constructed federated dataset encompasses eight distinct NLP tasks, culminating in a whole dataset comprising 2400 training examples and 1600 testing examples across all tasks. The specific tasks and datasets included in each federated dataset are cataloged in Table~\ref{table-dataset}. 

The NLP tasks within these datasets can be broadly divided into two types: generation tasks and classification tasks. To facilitate uniform processing by LLM, all tasks are converted into a generative format, employing distinct instructions for each dataset. Illustrative examples of these data for both classification and generation tasks are provided in Table~\ref{table-example}. For the input of the LLM, we adopted a simple template, the details of which are delineated in Table~\ref{table-template}.
\label{client_num}
\paragraph{Dataset Partitioning for Ablation Study.} In our ablation study in section~\ref{section:ablation} examining the client number to align with FL settings, we divided each task in our constructed federated datasets into five subsets, each comprising an equal number of training data. Based on our assumption that each client is associated with a single task, this division results in a total of 40 clients, with each client possessing a local dataset of 60 training examples. To mimic real-world FL communication dynamics, we employed a randomized selection process for clients (subsets) within each task according to specified sample rates. Accordingly, for sample rates specified as $\{0.2, 0.4, 0.6, 0.8, 1\}$, we selected 1,2,3,4, and 5 clients (subsets) per task, leading to 8, 16, 24, 32, and 40 clients participating in federated communications, respectively. The evaluation phase involves computing the average results across these selected clients for each specified sample rate, which provides a comprehensive analysis of how client numbers influence the performance of our method.

\paragraph{Different with Multi-Task Learning.} 
Although both multi-task learning and our test-time personalization setting involve multiple tasks during training and testing, multi-task learning operates in a centralized setting, whereas our setting is based on federated learning, a distributed setting. More importantly, our setting accounts for test-time distribution shifts, a challenge that is typically overlooked in conventional multi-task learning.
Additionally, fine-tuning on the combination of multi-task data from a centralized foundation model serves as a strong baseline for multi-task learning. In foundation models, all tasks are standardized into a uniform format, and the model benefits from task-agnostic token embeddings learned through extensive pre-training on diverse data. Thus, directly fine-tuning on this multi-task data represents the implementation of multi-task learning using foundation models ~\cite{yu2024unleashing}. We have included this baseline, referred to as "Centralized," in our experimental comparisons.

\begin{table}[h]
    \setlength{\tabcolsep}{4pt}
    \caption{Tasks and datasets of constructed federated dataset 1 and federated dataset 2.}
    \centering
    \begin{tabular}{ll|ll}
        \toprule[1.25pt]
         \multicolumn{2}{c|}{\bf Federated Dataset 1}& \multicolumn{2}{c}{\textbf{Federated Dataset 2}} \\
         Task & Dataset & Task & Dataset
\\ \hline 
Paraphrase & glue\_qqp & Paraphrase &paws\_wiki \\
Entailment & snli & Commonsense &hellaswag \\
Structure to text& web\_nlg\_en & Entailment &qnli \\
Text formatting & fix\_punct & Text formatting &word\_segment \\
Linguistic acceptability & cola & Summarization &gigaword \\
Word disambiguation & wic & Reading comprehension &bool\_q \\
Coreference & definite\_pronoun\_resolution & Sentiment &sentiment140 \\
Question classification & trec & Open-domain QA &acr\_easy \\
        \bottomrule[1.25pt]
    \end{tabular}
    \label{table-dataset}
\end{table}

\begin{table}
    \setlength{\tabcolsep}{4pt}
    \caption{Examples of data in our constructed federated datasets.}
    \centering
    \begin{tabular}{ll}
        \toprule[1.25pt]
         \multicolumn{2}{l}{\textbf{Data Examples}}  
\\ \hline 
\textbf{Input:} & \makecell[l]{The father convinced his son that it is possible for him to one day become a knight,\\but he may never achieve such status coming from a peasant family.\\ Who is "he"?\\ OPTIONS:\\- The father\\- his son}\\
\textbf{Output:} & His son \\
\hline
      \textbf{Input:} & \makecell[l]{Police are seeking a former village chief in north china for allegedly killing his \\political rivals in an attack apparently motivated by local power plays, state press \\reported monday .\\ Can you generate a short summary of the above paragraph?} \\
      \textbf{Output:} & Former chinese village head wanted for political murders \\
        \bottomrule[1.25pt]
    \end{tabular}
    \label{table-example}
\end{table}

\begin{table}
    \caption{Prompt Template.}
    \centering
    \begin{tabular}{ll}
        \toprule[1.25pt]
         &\multirow{1}{*}{\textbf{Template}}  
\\ \hline 
Prompt Input & \makecell[l]{\textbf{Instruction}: \{instruction\} \\  \textbf{Response}: }\\
      
        \bottomrule[1.25pt]
    \end{tabular}
    \label{table-template}
\end{table}

\subsection{Baselines and Implementation}
\label{baselines}
In this section, detailed descriptions of the implementation of FedDPA and each baseline compared in this study will be provided:

\begin{itemize}
\item \textbf{Centralized model:} This model is formulated by aggregating all available data from various tasks at a single centralized center for training purposes, with 50 epochs to optimize.

\item \textbf{Local-finetuned model:} This model trains independently without any external communication from other clients or a central server. It is specifically trained on data pertaining to a single task, dedicating 50 epochs to optimize for task-specific performance without the influence of external data.

\item \textbf{FedIT model \cite{zhang2023towards}:} The FedIT model is the final aggregated global model derived from diverse local client datasets after training. It embodies the essence of collaborative learning inherent to federated learning, assimilating knowledge from a multitude of client-specific data sources.

\item \textbf{FedLoRA model \cite{yi2023fedlora}:} Here, we adapt the training paradigm in paper \cite{yi2023fedlora} to NLP tasks by focusing on training the lightweight LoRA for aggregation while keeping the majority of the LLM parameters frozen. Subsequently, a personalized adaptation process is employed, where the globally aggregated LoRA undergoes further local training on each local client’s dataset to tailor the learning outcomes to individual client needs.

\item \textbf{FedDPA-F:} FedDPA-F is the combination of the global adapter and the local fine-tuning adapter. During the inference, the scale factor is set to $\lambda=1$ in the instance-wise dynamic weighting mechanism.

\item \textbf{FedDPA-T:} FedDPA-T is the combination of the global adapter and the local training adapter. Since the global adapter contributes to the training of the local adapter for personalization, it is essential to restrict the similarity score during the inference. This adjustment is necessary to ensure that the integration of the global and local adapters achieves optimal personalization outcomes. Thus, the scale factor $\lambda$ is set equal to the updating weight $\alpha$ used in the local adapter training.

\end{itemize}
All models are implemented using LoRA to enhance learning efficiency, with the rank of LoRA set as $r=8$ and only applied to $\mW_q$ and $\mW_v$. For FL methods, each client conducts 10 local epochs with a batch size of 32. We implement all the methods using PyTorch and conduct all experiments on NVIDIA Quadro RTX 6000 GPU.

\subsection{Algorithm}
\label{algorihtms}
\begin{algorithm}[H]
\caption{FedDPA-F}
\label{alg:linear}
\SetKwInOut{Input}{Require}
\SetKwInOut{Out}{Output}
\Input{Number of clients $M$; number of communication rounds $K$; local step size $\eta$; freeze foundation model $\vtheta$; initial global adapter $\Delta{\vtheta}_g$ and local adapters ${\Delta{\vtheta}_l^1,\cdots,\Delta{\vtheta}_l^M}$; local training datasets $\mD^1_{train},\cdots,\mD^M_{train}$.
}
\tcp{Learn Global Adapter}
\For{$k\leftarrow 1$ \KwTo $K$}{
{\textbf{Sample} clients $\mathcal{S}\subseteq\{1,\cdots,M\}$};\\
{\textbf{Communicate} $\Delta{\vtheta}_g$ to all clients $m\in \mathcal{S}$;}\\
\For{each client $m\in \mathcal{S}$ in parallel}{
{\textbf{Initialize} $\Delta{\vtheta}_g^m\leftarrow\Delta{\vtheta}_g$;}\\
{$\Delta{\vtheta}_g^m \leftarrow \textbf{Client local training}(\{\vtheta, \Delta{\vtheta}_g\}, \mD^m_{train}, \eta)$}; \hfill {[Equation \ref{eq-global}]} \\
{\textbf{Communicate} $\Delta{\vtheta}_g^m$ to the server;}\\
}
\textbf{Construct} $\Delta{\vtheta}_g =\sum_{m\in \mathcal{S}} \frac{1}{|\mathcal{S}|} \Delta{\vtheta}_g^m$; \hfill {[Equation \ref{eq-global}]}\\
}
\tcp{Learn Local Adapters}
\textcolor{red}{\For{each client $m\in \{1,\cdots,M\}$ in parallel}{
{\textbf{Initialize} $\Delta{\vtheta}_l^m\leftarrow\Delta{\vtheta}_g$;}\\
{$\Delta{\vtheta}_l^m \leftarrow \textbf{Client local training}(\{\vtheta, \Delta{\vtheta}_l^m\}, \mD^m_{train}, \eta)$}; \hfill {[Equation \ref{eq-local}]} \\
}}
{\textbf{return }$\Delta{\vtheta}_g, \{\Delta{\vtheta}_l^1,\cdots,\Delta{\vtheta}_l^M\}$.}
\end{algorithm}

\begin{algorithm}[H]
\caption{FedDPA-T}
\label{alg:FedDPA-T}
\SetKwInOut{Input}{Require}
\SetKwInOut{Out}{Output}
\Input{Number of clients $M$; number of communication rounds $K$; local step size $\eta$; freeze foundation model $\vtheta$; initial global adapter $\Delta{\vtheta}_g$ and local adapters ${\Delta{\vtheta}_l^1,\cdots,\Delta{\vtheta}_l^M}$; local training datasets $\mD^1_{train},\cdots,\mD^M_{train}$.
}
\For{$k\leftarrow 1$ \KwTo $K$}{
{\textbf{Sample} clients $\mathcal{S}\subseteq\{1,\cdots,M\}$};\\
{\textbf{Communicate} $\Delta{\vtheta}_g$ to all clients $m\in \mathcal{S}$;}\\
\For{each client $m\in \mathcal{S}$ in parallel}{
{\textbf{Initialize} $\Delta{\vtheta}_g^m\leftarrow\Delta{\vtheta}_g$ and \textcolor{red}{$\Delta{\vtheta}_l^{m}\leftarrow\Delta{\vtheta}_l^{m,k-1}$}; }\\
\tcp{Learn Global Adapter} 
{$\Delta{\vtheta}_g^m \leftarrow \textbf{Client local training}(\{\vtheta, \Delta{\vtheta}_g\}, \mD^m_{train}, \eta)$}; \hfill {[Equation \ref{eq-global}]} \\
{\tcp{Learn Local Adapter}} 
\textcolor{red}{$\Delta{\vtheta}_l^{m,k} \leftarrow \textbf{Client local training}(\{\vtheta, \Delta{\vtheta}_g,\Delta{\vtheta}_l^m\}, \mD^m_{train}, \eta)$}; \hfill {[Equation \ref{eq-local}]} \\
{\textbf{Communicate} $\Delta{\vtheta}_g^m$ to the server;}\\
\textcolor{red}{\textbf{Maintain} $\Delta{\vtheta}_l^{m,k}$ locally;}\\

}
\textbf{Construct} $\Delta{\vtheta}_g =\sum_{m\in \mathcal{S}} \frac{1}{|\mathcal{S}|} \Delta{\vtheta}_g^m$; \hfill {[Equation \ref{eq-global}]}\\
} 
{\textbf{return }$\Delta{\vtheta}_g, \{\Delta{\vtheta}_l^1,\cdots,\Delta{\vtheta}_l^M\}$.}
\end{algorithm}

\section{Additional Experiments}

\subsection{Adaptability Analysis}
\label{adapt}
To enhance applicability across diverse non-IID environments, our method is meticulously designed with a high degree of flexibility for its adaption across various global learning frameworks, backbones and PEFT methods for different scenarios. This adaptability is simply achieved through the straightforward substitution of the FedAvg, LLM and LoRA with alternative aggregation methods, transformer-based foundation models and adapter-based PEFT methods during the training. In our experiment, we employ FedAvg, LLM and LoRA as representative examples, demonstrating our methods' superior performance compared to other baselines as indicated in Table~\ref{table-best} and Table~\ref{table-data2}. To further validate the effectiveness and versatility of our approach within different federated learning contexts, we adapt our methods to include the FedProx\cite{li2020federated} framework and also implement other baselines (FedIT and FedLoRA) within FedProx to maintain a fair comparison.

Results presented in Table~\ref{table-fedprox} indicate that our methods, both training and fine-tuning methods, outperform competing approaches. Specifically, FedDPA-T excels in personalization, while FedDPA-F leads in test-time personalization, maintaining consistent performance with FedAvg. These findings underscore the robustness and consistent efficacy of our methods across various global learning paradigms for different non-IID scenarios. 
\begin{table}[h]
    \small
    \setlength{\tabcolsep}{4pt}
    \caption{Personalization and test-time personalization results of different models with FedProx framework on federated dataset 1. FedDPA-F represents the model with the local fine-tuning adapter and FedDPA-T represents the model with the local training adapter. Linguistic represents the linguistic acceptability task, Word Dis represents the word disambiguation task, and Question CLS represents question classification task.}
    \centering
    \begin{tabular}{l|cccccccc|c}
        \toprule[1.25pt]
        \multirow{2}{*}{Methods}  & \multicolumn{9}{c}{\bf Federated Dataset 1} \\
        &\multicolumn{1}{l}{\makecell[c]{Para \\ -phrase}}  &\multicolumn{1}{l}{\makecell[c]{Entail \\ -ment}}  &\multicolumn{1}{l}{\makecell[c]{Structure \\ to Text}} &\multicolumn{1}{l}{\makecell[c]{Text For \\ -matting}} &\multicolumn{1}{l}{\makecell[c]{Linguistic \\ Acc}}  &\multicolumn{1}{l}{\makecell[c]{Word \\ Dis}} &\multicolumn{1}{l}{\makecell[c]{Core \\ -ference}} &\multicolumn{1}{l}{\makecell[c]{Question \\ CLS}} &\multicolumn{1}{l}{\makecell[c]{Average}}
\\ \hline 
\rowcolor{mygray}\multicolumn{10}{l}{ \emph{Personalization}} \\
FedIT         &71.00  &83.00  &70.38  &96.32  &69.50 &63.50 &73.55 &91.50 &77.34 \\
FedLoRA       &79.00  &85.00  &70.63  &96.60  &69.00 &63.00 &\textbf{78.85} &88.50 &78.82 \\
\hline
FedDPA-F &77.50 &85.00  &\textbf{71.92}  &96.78  &73.00 &\textbf{63.50} &77.87 &89.50 &79.38 \\
\rowcolor{aliceblue!60} FedDPA-T     &\textbf{79.50}  &\textbf{85.50}  &71.62  &\textbf{96.89}  &\textbf{76.00} &60.00 &77.28 &\textbf{93.50} &\textbf{80.04}\\

\hline 
 \rowcolor{mygray}\multicolumn{10}{l}{ \emph{Test-Time Personalization}} \\
FedLoRA       &76.76  &75.82  &74.11 &74.36  &75.01 &72.27 &77.16 &74.87 &75.05 \\
\hline 
\rowcolor{aliceblue!60} FedDPA-F &\textbf{77.25} &\textbf{76.01}  &\textbf{76.72}  &\textbf{76.95}  &\textbf{77.21} &\textbf{75.48} &\textbf{77.31} &\textbf{76.14} &\textbf{76.63} \\
FedDPA-T     &75.64  &74.62  &75.58  &75.01  &74.90 &74.21 &74.80 &75.35 &75.01\\
        \bottomrule[1.25pt]
    \end{tabular}
    \label{table-fedprox}
\end{table}

\subsection{Instance-Wise Dynamic Weighting Mechanism Analysis}
\label{auto-exp}
In this section, we further examine the impact of the instance-wise dynamic weighting mechanism, including the similarity metric, the selected local instance number and the type of instance representation.

\paragraph{Impact of Similarity Metric.} 
In Section~\ref{auto}, we employ the similarity metric to calculate the average similarity scores of each input instance, which serves as the weight $\alpha_t$ to dynamically balance the global and local adapters. For this purpose, cosine similarity is selected in our experiment due to its better robustness and normalization with high-dimensional vectors than other metrics, and its superiority has been demonstrated in many NLP/CV works\cite{radford2021learning,yang2024dgl,yang2024eva}. Additionally, we conducted an ablation study comparing other metrics like L2-norm and Pearson correlation, and the results in Table~\ref{table-metric} demonstrate that cosine similarity outperforms other similarity metrics.

\begin{figure}[h]
\begin{center}
\includegraphics[width=0.7\textwidth]{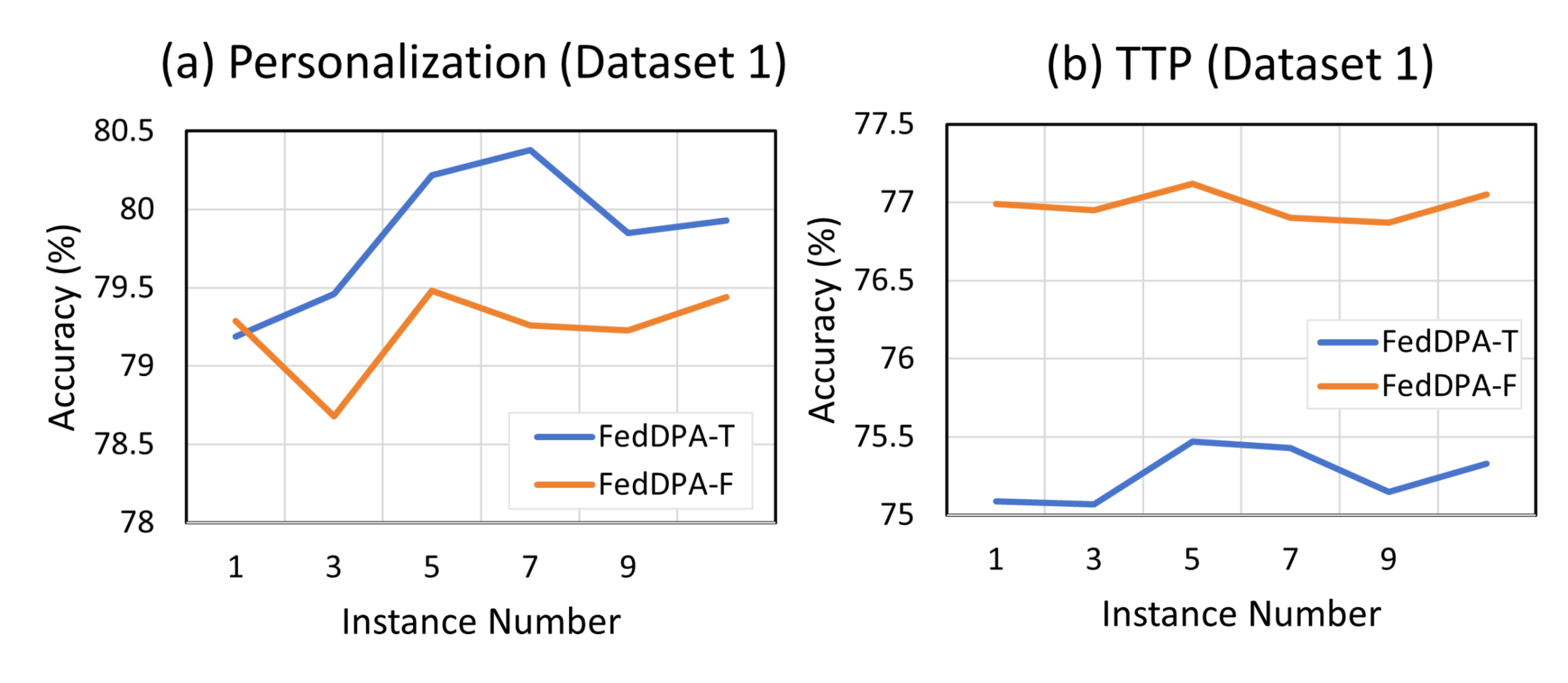} 
\end{center}
\caption{Average accuracy varies as different instance numbers. TTP represents test-time personalization.}
\label{fig:instancenum}
\end{figure}

\paragraph{Impact of Instance Number $S$.} In Section~\ref{auto}, the selection of $S$, representing the number of local instances for similarity calculation, is pivotal. To comprehensively evaluate the effect of varying the number of these instances, we conduct a series of experiments employing distinct local instance numbers, specifically $ S \in \{1,3,5,7,9\}$. The accuracy results, as depicted in Figure~\ref{fig:instancenum}, illustrate the dependency of model performance on different instance numbers $S$. 
As demonstrated in Figure~\ref{fig:instancenum} (a), in the context of personalization, it is observed that our models attain a plateau in accuracy when the instance number exceeds 5. This indicates a stabilization in model performance beyond this threshold of local instances. Furthermore, Figure~\ref{fig:instancenum} (b) delves into the realm of test-time personalization. The findings here reveal similar results, indicating that variations in the instance number do not markedly impact the model’s performance in test-time personalization. 

\paragraph{Impact of Instance Representation.} 
In Section~\ref{auto}, our method entails utilizing the embedding of the final token from the last hidden layer of the LLM, denoted as 'LAST', as the input instance representation for the purpose of similarity calculation. In this exploration, we delve into another instance representation strategy, which involves employing the average embedding of all tokens from the final hidden layer of the LLM, herein referred to as 'AVG'. The comparative analysis, as presented in Table~\ref{table-representation}, demonstrates that employing the embedding of the last token yields superior performance relative to the strategy of averaging the embeddings of all tokens. This observed difference in performance can be attributed to the decoder structure inherent to LLMs, wherein the final token is capable of attending to all preceding tokens, thereby encapsulating comprehensive sentence-level information.
\begin{minipage}{\textwidth}
\begin{minipage}[t]{0.48\textwidth}
    \makeatletter\def\@captype{table}
    \caption{Ablation study of similarity metric (Sim). P represents personalization, and TTP represents test-time personalization. -L2 represents using the L2-Norm as metric, Pearson represents using the Pearson correlation as metric, and Cosine represents using the cosine similarity as the metric.}
    \centering
    \begin{tabular}{ll|ll}
        \toprule[1.25pt]
        \multirow{2}{*}{{Methods}} &\multirow{2}{*}{{Sim}} &\multicolumn{2}{c}{\bf Fed Dataset 1} \\
        & & \multicolumn{1}{c}{P} & \multicolumn{1}{c}{TTP}
\\ \hline 
\multirow{2}{*}{\makecell[c]{FedDPA-F}}&\multicolumn{1}{l|}{-L2} &77.69 &77.10 \\
&\multicolumn{1}{l|}{Pearson} &79.09 &76.78 \\
&\multicolumn{1}{l|}{Cosine} &\textbf{79.48} &\textbf{77.12} \\
\hline
\multirow{2}{*}{\makecell[c]{FedDPA-T}}&\multicolumn{1}{l|}{-L2} &77.58 &\textbf{77.30} \\
&\multicolumn{1}{l|}{Pearson} &79.73 &75.27 \\
&\multicolumn{1}{l|}{Cosine} &\textbf{80.22} &75.47 \\
      
        \bottomrule[1.25pt]
    \end{tabular}
    \label{table-metric}
\end{minipage}
\hfill
\begin{minipage}[t]{0.48\textwidth}
    \makeatletter\def\@captype{table}
    \caption{Ablation study of instance representations (Emb). P represents personalization, and TTP represents test-time personalization. LAST represents using the embedding of the final token from the final hidden layer of LLM as instance representation, and AVG represents using the average embedding of all tokens from the final hidden layer of LLM as instance representation.}
    \centering
    \begin{tabular}{ll|ll}
        \toprule[1.25pt]
        \multirow{2}{*}{{Methods}} &\multirow{2}{*}{{Emb}} &\multicolumn{2}{c}{\bf Fed Dataset 1} \\
        & & \multicolumn{1}{c}{P} & \multicolumn{1}{c}{TTP}
\\ \hline 
\multirow{2}{*}{\makecell[c]{FedDPA-F}}&\multicolumn{1}{l|}{AVG} &79.30 &76.77 \\
&\multicolumn{1}{l|}{LAST} &\textbf{79.48} &\textbf{77.12} \\
\hline
\multirow{2}{*}{\makecell[c]{FedDPA-T}}&\multicolumn{1}{l|}{AVG} &79.65 &73.36 \\
&\multicolumn{1}{l|}{LAST} &\textbf{80.22} &\textbf{75.47} \\
      
        \bottomrule[1.25pt]
    \end{tabular}
    \label{table-representation}
\end{minipage}
\end{minipage}

\subsection{Model Scalability Analysis}
\label{scale}
In order to examine the effectiveness of model scalability, we conduct experiments based on a larger model, LLaMA-13B. 
The outcomes, as presented in Table~\ref{table-size}, elucidate that larger models exhibit superior performance over their smaller counterparts across all personalization methods evaluated. Furthermore, it is noteworthy that FedDPA-T surpasses FedDPA-F in terms of personalization and achieves comparable results in test-time personalization. This analysis underscores the inherent advantages of larger models in enhancing model performance, alongside the advance of the FedDPA-T approach in the context of personalization and adaptability to test-time conditions.
\begin{table}[h]
\centering
\caption{Ablation study of model size. P represents personalization, and TTP represents test-time personalization.}
    \begin{tabular}{ll|ll}
        \toprule[1.25pt]
        \multirow{2}{*}{{Methods}} &\multirow{2}{*}{{Size}} &\multicolumn{2}{c}{\bf Fed Dataset 1} \\
        & & \multicolumn{1}{c}{P} & \multicolumn{1}{c}{TTP}
\\ \hline 
\multirow{2}{*}{\makecell[c]{FedDPA-F}}&\multicolumn{1}{l|}{7B} &79.48 &77.12 \\
&\multicolumn{1}{l|}{13B} &\textbf{81.52} &\textbf{80.55} \\
\hline
\multirow{2}{*}{\makecell[c]{FedDPA-T}}&\multicolumn{1}{l|}{7B} &80.22 &75.47 \\
&\multicolumn{1}{l|}{13B} &\textbf{82.76} &\textbf{80.47} \\
      
        \bottomrule[1.25pt]
    \end{tabular}
    \label{table-size}
\end{table}

\subsection{Communication and Computation Analysis}
\label{computation}
In this section, we undertake a detailed examination of both the communication and computation overhead associated with our proposed model in comparison to other baseline models. The results, as detailed in Table~\ref{table-com}, delineate the communication and computation burdens imposed by various models. Given that these models are all based on the LoRA framework and exclusively transmit LoRA weights for aggregation (with our methods specifically transmitting only the global LoRA weights), they inherently sustain a minimal communication overhead. Regarding the computation overhead, the LoRA architecture permits the training of both local and global LoRAs in parallel, resulting in a marginal increase in computational demands for FedDPA-T. Conversely, FedDPA-F learns the local LoRA through an additional fine-tuning phase, thereby not imposing any additional computational overhead during the training phase. 

Additionally, we have conducted an analysis of the inference time associated with our models. This examination involved calculating the average inference time per instance for FedLoRA, FedDPA without the instance-wise dynamic weighting mechanism, and FedDPA. As illustrated in Table~\ref{table-infer}, it is observed that our methods incur slightly higher inference time compared to FedLoRA. This marginal increase in inference time underscores the efficiency of our proposed methods, demonstrating that the enhanced performance and capabilities are achieved with a minimal impact on computational efficiency during inference.
\begin{table}[h]
\begin{minipage}[t]{0.5\textwidth}
 \centering
 \caption{The communication and computation overhead of FedDPA and other baselines on Federated Dataset 1.}
    \begin{tabular}{l|ll}
        \toprule[1.25pt]
        \multirow{1}{*}{{Methods}} &\multicolumn{1}{c}{Comm.Overhead} & \multicolumn{1}{c}{Comp.Overhead}
\\ \hline 
FedIT&4.2M(0.06\%)  &0.277 TFLOPS \\
FedLoRA &4.2M(0.06\%) &0.277 TFLOPS \\
\hline
FedDPA-F &4.2M(0.06\%) &0.277 TFLOPS \\
FedDPA-T &4.2M(0.06\%) &0.281 TFLOPS \\
        \bottomrule[1.25pt]
    \end{tabular}
    \label{table-com}
\end{minipage}
\hfill
\begin{minipage}[t]{0.4\textwidth}
\centering
\caption{Average inference time per instance. Auto represents the instance-wise dynamic weighting mechanism. }
    \begin{tabular}{l|l}
        \toprule[1.25pt]
        \multirow{1}{*}{{Methods}} &\multicolumn{1}{c}{Time} 
\\ \hline 
FedLoRA&3.84s \\
FedDPA (w/o auto) &3.91s \\
FedDPA &4.13s \\
        \bottomrule[1.25pt]
    \end{tabular}
    \label{table-infer}
\end{minipage}
\end{table}

\section{Discussions}
\label{limit}
\subsection{Extend to Other Scenarios of Test-Time FL}
In this paper, we primarily consider an ideal scenario for our proposed test-time personalization setting, where all tasks are included for all clients. In this section, we will discuss our methods under alternative scenarios.

In practical applications, it is possible that some tasks remain unseen by all clients during training and may only appear during the testing phase for certain clients. In this scenario, test-time distribution shifts arise from these unseen tasks. According to previous works \cite{charles2024towards,tan2024heterogeneity}, generic features learned through FL are robust to distribution shifts, even those originating from unseen test-time tasks. Consequently, our method can be directly adapted to this scenario. We conducted experiments to evaluate our methods against other baselines on three unseen test-time tasks. All methods and baselines are trained on our constructed Federated Dataset 1 and tested on three tasks not included in Federated Dataset 1. For personalized methods, we report the average score across all clients and the maximum score among all clients to provide a comprehensive comparison.

As shown in Table~\ref{table-other}, FedDPA-T outperforms all other models, indicating the effectiveness of our method on unseen test-time tasks. Additionally, FedDPA-F surpasses FedLoRA, suggesting that the generic features learned through FL across diverse data distributions are robust to various distribution shifts, consistent with the findings in \cite{tan2024heterogeneity}. Despite this, other techniques targeting these unseen test-time tasks could further enhance our proposed methods, which we leave for future work to explore.

\begin{table}[h]
\centering
\caption{Test-time performance on unseen tasks. All models are trained on Federated Dataset 1, and these unseen test-time tasks are not included in Federated Dataset 1. AVG represents the average score across all clients for this task, while MAX represents the highest score among these clients for this task. The best performance for AVG is bolded, and the best performance for MAX is underlined.  }
    \begin{tabular}{llll|llll}
        \toprule[1.25pt]
        \multirow{2}{*}{{Test-Time Task}} &\multirow{2}{*}{FedIT} & \multicolumn{2}{c}{FedLoRA} & \multicolumn{2}{c}{FedDPA-F} & \multicolumn{2}{c}{FedDPA-T} \\
        & & AVG & MAX & AVG & MAX & AVG & MAX
\\ \hline 
Summarization & 22.40 & 22.25 & 22.42 & 22.37 & 22.52 & \textbf{22.46} & \underline{22.65}\\
Reading Comprehension & 69.50 & 68.88 & 73.00 & 69.44 & 72.00 & \textbf{71.88} & \underline{76.00} \\
Open Domain QA & 78.32 & 76.46 & 78.96 & 78.01 & 79.61 & \textbf{78.76} & \underline{79.92} \\
        \bottomrule[1.25pt]
    \end{tabular}
    \label{table-other}
\end{table}

Our methods are also applicable to scenarios involving the introduction of new clients. As previously analyzed, our methods are robust to different distribution shifts. By computing the similarity between instances from new clients and existing clients through our instance-wise dynamic weighting mechanism, we can identify the most similar existing client. The model of this identified client can then be used as the initial model for the new clients, providing a more effective starting point for further training.

\subsection{Limitations}
The proposed dual-personalizing adapter architecture is limited by 1) model scales: the proposed methods rely on the foundation model, presupposing that each client possesses sufficient memory capacity and computational resources to store and train the foundation model with PEFT methods. 2) secure issues: the framework operates under the assumption that all clients are trusted and legally entitled to access and utilize data stored on them, and the whole process does not suffer from any attacks.

\end{document}

%% file: math_commands.tex
\usepackage{amsmath,amsfonts,bm}

\def\eqref#1{equation~\ref{#1}}

\def\1{\bm{1}}

\def\vtheta{{\bm{\theta}}}

\def\vh{{\bm{h}}}

\def\vw{{\bm{w}}}
\def\vx{{\bm{x}}}
\def\vy{{\bm{y}}}

\def\mD{{\bm{D}}}

\def\mW{{\bm{W}}}

\DeclareMathAlphabet{\mathsfit}{\encodingdefault}{\sfdefault}{m}{sl}
\SetMathAlphabet{\mathsfit}{bold}{\encodingdefault}{\sfdefault}{bx}{n}

\def\sD{{\mathbb{D}}}

\newcommand{\Ls}{\mathcal{L}}

\DeclareMathOperator*{\argmin}{arg\,min}

%% file: neurips_2024.bbl
\begin{thebibliography}{10}

\bibitem{arjovsky2020out}
Martin Arjovsky.
\newblock {\em Out of distribution generalization in machine learning}.
\newblock PhD thesis, New York University, 2020.

\bibitem{babakniya2023slora}
Sara Babakniya, Ahmed~Roushdy Elkordy, Yahya~H Ezzeldin, Qingfeng Liu, Kee-Bong Song, Mostafa El-Khamy, and Salman Avestimehr.
\newblock Slora: Federated parameter efficient fine-tuning of language models.
\newblock {\em arXiv preprint arXiv:2308.06522}, 2023.

\bibitem{bao2024adaptive}
Wenxuan Bao, Tianxin Wei, Haohan Wang, and Jingrui He.
\newblock Adaptive test-time personalization for federated learning.
\newblock {\em Advances in Neural Information Processing Systems}, 36, 2024.

\bibitem{brown2020language}
Tom Brown, Benjamin Mann, Nick Ryder, Melanie Subbiah, Jared~D Kaplan, Prafulla Dhariwal, Arvind Neelakantan, Pranav Shyam, Girish Sastry, Amanda Askell, et~al.
\newblock Language models are few-shot learners.
\newblock {\em Advances in neural information processing systems}, 33:1877--1901, 2020.

\bibitem{charles2024towards}
Zachary Charles, Nicole Mitchell, Krishna Pillutla, Michael Reneer, and Zachary Garrett.
\newblock Towards federated foundation models: Scalable dataset pipelines for group-structured learning.
\newblock {\em Advances in Neural Information Processing Systems}, 36, 2024.

\bibitem{chen2024feddat}
Haokun Chen, Yao Zhang, Denis Krompass, Jindong Gu, and Volker Tresp.
\newblock Feddat: An approach for foundation model finetuning in multi-modal heterogeneous federated learning.
\newblock In {\em Proceedings of the AAAI Conference on Artificial Intelligence}, volume~38, pages 11285--11293, 2024.

\bibitem{collins2021exploiting}
Liam Collins, Hamed Hassani, Aryan Mokhtari, and Sanjay Shakkottai.
\newblock Exploiting shared representations for personalized federated learning.
\newblock In {\em International conference on machine learning}, pages 2089--2099. PMLR, 2021.

\bibitem{dong2022federated}
Jiahua Dong, Lixu Wang, Zhen Fang, Gan Sun, Shichao Xu, Xiao Wang, and Qi~Zhu.
\newblock Federated class-incremental learning.
\newblock In {\em Proceedings of the IEEE/CVF conference on computer vision and pattern recognition}, pages 10164--10173, 2022.

\bibitem{edalati2022krona}
Ali Edalati, Marzieh Tahaei, Ivan Kobyzev, Vahid~Partovi Nia, James~J Clark, and Mehdi Rezagholizadeh.
\newblock Krona: Parameter efficient tuning with kronecker adapter.
\newblock {\em arXiv preprint arXiv:2212.10650}, 2022.

\bibitem{fallah2020personalized}
Alireza Fallah, Aryan Mokhtari, and Asuman Ozdaglar.
\newblock Personalized federated learning with theoretical guarantees: A model-agnostic meta-learning approach.
\newblock {\em Advances in Neural Information Processing Systems}, 33:3557--3568, 2020.

\bibitem{he2021towards}
Junxian He, Chunting Zhou, Xuezhe Ma, Taylor Berg-Kirkpatrick, and Graham Neubig.
\newblock Towards a unified view of parameter-efficient transfer learning.
\newblock {\em arXiv preprint arXiv:2110.04366}, 2021.

\bibitem{houlsby2019parameter}
Neil Houlsby, Andrei Giurgiu, Stanislaw Jastrzebski, Bruna Morrone, Quentin De~Laroussilhe, Andrea Gesmundo, Mona Attariyan, and Sylvain Gelly.
\newblock Parameter-efficient transfer learning for nlp.
\newblock In {\em International Conference on Machine Learning}, pages 2790--2799. PMLR, 2019.

\bibitem{hu2021lora}
Edward~J Hu, Yelong Shen, Phillip Wallis, Zeyuan Allen-Zhu, Yuanzhi Li, Shean Wang, Lu~Wang, and Weizhu Chen.
\newblock Lora: Low-rank adaptation of large language models.
\newblock {\em arXiv preprint arXiv:2106.09685}, 2021.

\bibitem{hu2023llm}
Zhiqiang Hu, Yihuai Lan, Lei Wang, Wanyu Xu, Ee-Peng Lim, Roy Ka-Wei Lee, Lidong Bing, and Soujanya Poria.
\newblock Llm-adapters: An adapter family for parameter-efficient fine-tuning of large language models.
\newblock {\em arXiv preprint arXiv:2304.01933}, 2023.

\bibitem{jiang2023low}
Jingang Jiang, Xiangyang Liu, and Chenyou Fan.
\newblock Low-parameter federated learning with large language models.
\newblock {\em arXiv preprint arXiv:2307.13896}, 2023.

\bibitem{jiang2023test}
Liangze Jiang and Tao Lin.
\newblock Test-time robust personalization for federated learning.
\newblock In {\em ICLR}, 2023.

\bibitem{kuang2023federatedscope}
Weirui Kuang, Bingchen Qian, Zitao Li, Daoyuan Chen, Dawei Gao, Xuchen Pan, Yuexiang Xie, Yaliang Li, Bolin Ding, and Jingren Zhou.
\newblock Federatedscope-llm: A comprehensive package for fine-tuning large language models in federated learning.
\newblock {\em arXiv preprint arXiv:2309.00363}, 2023.

\bibitem{lester2021power}
Brian Lester, Rami Al-Rfou, and Noah Constant.
\newblock The power of scale for parameter-efficient prompt tuning.
\newblock {\em arXiv preprint arXiv:2104.08691}, 2021.

\bibitem{li2021ditto}
Tian Li, Shengyuan Hu, Ahmad Beirami, and Virginia Smith.
\newblock Ditto: Fair and robust federated learning through personalization.
\newblock In {\em International Conference on Machine Learning}, pages 6357--6368. PMLR, 2021.

\bibitem{li2020federated}
Tian Li, Anit~Kumar Sahu, Manzil Zaheer, Maziar Sanjabi, Ameet Talwalkar, and Virginia Smith.
\newblock Federated optimization in heterogeneous networks.
\newblock {\em Proceedings of Machine learning and systems}, 2:429--450, 2020.

\bibitem{li2021prefix}
Xiang~Lisa Li and Percy Liang.
\newblock Prefix-tuning: Optimizing continuous prompts for generation.
\newblock {\em arXiv preprint arXiv:2101.00190}, 2021.

\bibitem{li2021fedbn}
Xiaoxiao Li, Meirui Jiang, Xiaofei Zhang, Michael Kamp, and Qi~Dou.
\newblock Fedbn: Federated learning on non-iid features via local batch normalization.
\newblock {\em arXiv preprint arXiv:2102.07623}, 2021.

\bibitem{masoudnia2014mixture}
Saeed Masoudnia and Reza Ebrahimpour.
\newblock Mixture of experts: a literature survey.
\newblock {\em Artificial Intelligence Review}, 42:275--293, 2014.

\bibitem{mcmahan2017communication}
Brendan McMahan, Eider Moore, Daniel Ramage, Seth Hampson, and Blaise~Aguera y~Arcas.
\newblock Communication-efficient learning of deep networks from decentralized data.
\newblock In {\em Artificial intelligence and statistics}, pages 1273--1282. PMLR, 2017.

\bibitem{radford2021learning}
Alec Radford, Jong~Wook Kim, Chris Hallacy, Aditya Ramesh, Gabriel Goh, Sandhini Agarwal, Girish Sastry, Amanda Askell, Pamela Mishkin, Jack Clark, et~al.
\newblock Learning transferable visual models from natural language supervision.
\newblock In {\em International conference on machine learning}, pages 8748--8763. PMLR, 2021.

\bibitem{ren2024advances}
Chao Ren, Han Yu, Hongyi Peng, Xiaoli Tang, Anran Li, Yulan Gao, Alysa~Ziying Tan, Bo~Zhao, Xiaoxiao Li, Zengxiang Li, et~al.
\newblock Advances and open challenges in federated learning with foundation models.
\newblock {\em arXiv preprint arXiv:2404.15381}, 2024.

\bibitem{sun2023fedbpt}
Jingwei Sun, Ziyue Xu, Hongxu Yin, Dong Yang, Daguang Xu, Yiran Chen, and Holger~R Roth.
\newblock Fedbpt: Efficient federated black-box prompt tuning for large language models.
\newblock {\em arXiv preprint arXiv:2310.01467}, 2023.

\bibitem{tan2022towards}
Alysa~Ziying Tan, Han Yu, Lizhen Cui, and Qiang Yang.
\newblock Towards personalized federated learning.
\newblock {\em IEEE Transactions on Neural Networks and Learning Systems}, 2022.

\bibitem{tan2024heterogeneity}
Yue Tan, Chen Chen, Weiming Zhuang, Xin Dong, Lingjuan Lyu, and Guodong Long.
\newblock Is heterogeneity notorious? taming heterogeneity to handle test-time shift in federated learning.
\newblock {\em Advances in Neural Information Processing Systems}, 36, 2024.

\bibitem{wang2023far}
Yizhong Wang, Hamish Ivison, Pradeep Dasigi, Jack Hessel, Tushar Khot, Khyathi~Raghavi Chandu, David Wadden, Kelsey MacMillan, Noah~A Smith, Iz~Beltagy, et~al.
\newblock How far can camels go? exploring the state of instruction tuning on open resources.
\newblock {\em arXiv preprint arXiv:2306.04751}, 2023.

\bibitem{wei2021finetuned}
Jason Wei, Maarten Bosma, Vincent~Y Zhao, Kelvin Guu, Adams~Wei Yu, Brian Lester, Nan Du, Andrew~M Dai, and Quoc~V Le.
\newblock Finetuned language models are zero-shot learners.
\newblock {\em arXiv preprint arXiv:2109.01652}, 2021.

\bibitem{xu2023joint}
Jian Xu and Shao-Lun Huang.
\newblock A joint training-calibration framework for test-time personalization with label shift in federated learning.
\newblock In {\em Proceedings of the 32nd ACM International Conference on Information and Knowledge Management}, pages 4370--4374, 2023.

\bibitem{xu2023parameter}
Lingling Xu, Haoran Xie, Si-Zhao~Joe Qin, Xiaohui Tao, and Fu~Lee Wang.
\newblock Parameter-efficient fine-tuning methods for pretrained language models: A critical review and assessment.
\newblock {\em arXiv preprint arXiv:2312.12148}, 2023.

\bibitem{xu2024fwdllm}
M~Xu, D~Cai, Y~Wu, X~Li, and S~Wang.
\newblock Fwdllm: Efficient fedllm using forward gradient.
\newblock 2024.

\bibitem{xu2023federated}
Mengwei Xu, Yaozong Wu, Dongqi Cai, Xiang Li, and Shangguang Wang.
\newblock Federated fine-tuning of billion-sized language models across mobile devices.
\newblock {\em arXiv preprint arXiv:2308.13894}, 2023.

\bibitem{yang2024eva}
Xiangpeng Yang, Linchao Zhu, Hehe Fan, and Yi~Yang.
\newblock Eva: Zero-shot accurate attributes and multi-object video editing.
\newblock {\em arXiv preprint arXiv:2403.16111}, 2024.

\bibitem{yang2024dgl}
Xiangpeng Yang, Linchao Zhu, Xiaohan Wang, and Yi~Yang.
\newblock Dgl: Dynamic global-local prompt tuning for text-video retrieval.
\newblock In {\em Proceedings of the AAAI Conference on Artificial Intelligence}, volume~38, pages 6540--6548, 2024.

\bibitem{yi2023fedlora}
Liping Yi, Han Yu, Gang Wang, and Xiaoguang Liu.
\newblock Fedlora: Model-heterogeneous personalized federated learning with lora tuning.
\newblock {\em arXiv preprint arXiv:2310.13283}, 2023.

\bibitem{yoon2021federated}
Jaehong Yoon, Wonyong Jeong, Giwoong Lee, Eunho Yang, and Sung~Ju Hwang.
\newblock Federated continual learning with weighted inter-client transfer.
\newblock In {\em International Conference on Machine Learning}, pages 12073--12086. PMLR, 2021.

\bibitem{yu2024unleashing}
Jun Yu, Yutong Dai, Xiaokang Liu, Jin Huang, Yishan Shen, Ke~Zhang, Rong Zhou, Eashan Adhikarla, Wenxuan Ye, Yixin Liu, et~al.
\newblock Unleashing the power of multi-task learning: A comprehensive survey spanning traditional, deep, and pretrained foundation model eras.
\newblock {\em arXiv preprint arXiv:2404.18961}, 2024.

\bibitem{yu2023federated}
Sixing Yu, J~Pablo Mu{\~n}oz, and Ali Jannesari.
\newblock Federated foundation models: Privacy-preserving and collaborative learning for large models.
\newblock {\em arXiv preprint arXiv:2305.11414}, 2023.

\bibitem{zhang2023towards}
Jianyi Zhang, Saeed Vahidian, Martin Kuo, Chunyuan Li, Ruiyi Zhang, Guoyin Wang, and Yiran Chen.
\newblock Towards building the federated gpt: Federated instruction tuning.
\newblock {\em arXiv preprint arXiv:2305.05644}, 2023.

\bibitem{zhang2023fedpetuning}
Zhuo Zhang, Yuanhang Yang, Yong Dai, Qifan Wang, Yue Yu, Lizhen Qu, and Zenglin Xu.
\newblock Fedpetuning: When federated learning meets the parameter-efficient tuning methods of pre-trained language models.
\newblock In {\em Annual Meeting of the Association of Computational Linguistics 2023}, pages 9963--9977. Association for Computational Linguistics (ACL), 2023.

\bibitem{zhuang2023foundation}
Weiming Zhuang, Chen Chen, and Lingjuan Lyu.
\newblock When foundation model meets federated learning: Motivations, challenges, and future directions.
\newblock {\em arXiv preprint arXiv:2306.15546}, 2023.

\end{thebibliography}
